\documentclass{article}

\usepackage{microtype}
\usepackage{graphicx}
\usepackage{subfigure}
\usepackage{booktabs} % for professional tables

\usepackage{hyperref}

\usepackage[accepted]{assets/icml2023}

\usepackage{amsmath}
\usepackage{amssymb}
\usepackage{mathtools}
\usepackage{amsthm}

\usepackage{amsmath,amssymb,amsfonts}
\usepackage{bm}
\usepackage{pifont}
\usepackage{mathrsfs}

\usepackage[ruled]{algorithm2e}

\usepackage{graphicx}
\usepackage{textcomp}

\usepackage{mathtools}
\usepackage{multirow}
\usepackage{makecell}
\usepackage{setspace}

\usepackage{booktabs}
\usepackage{colortbl}

\usepackage{url}            % simple URL typesetting
\usepackage{nicefrac}       % compact symbols for 1/2, etc.
\usepackage{microtype}      % microtypography
\usepackage{xcolor}         % colors
\usepackage{bbm}

\usepackage{dsfont} % indicator function

\usepackage{float,wrapfig}

\usepackage{array}
\usepackage{adjustbox}
\usepackage{multirow}
\usepackage{colortbl}
\usepackage{caption}
\usepackage{amsmath}
\usepackage{multicol}
\usepackage{mathtools}

\usepackage{soul}
\usepackage{enumitem}
\usepackage{xspace}

\makeatletter
\DeclareRobustCommand\onedot{\futurelet\@let@token\@onedot}
\def\@onedot{\ifx\@let@token.\else.\null\fi\xspace}

\def\eg{e.g\onedot}

\makeatother

\newcommand{\ignore}[1]{}

\def\numpairs{26\xspace}

\DeclarePairedDelimiter\abs{\lvert}{\rvert}
\DeclarePairedDelimiter\norm{\lVert}{\rVert}
\makeatletter
\let\oldabs\abs
\def\abs{\@ifstar{\oldabs}{\oldabs*}}
\let\oldnorm\norm
\def\norm{\@ifstar{\oldnorm}{\oldnorm*}}
\makeatother

\def\eqref#1{equation~\ref{#1}}

\DeclareMathAlphabet{\mathsfit}{\encodingdefault}{\sfdefault}{m}{sl}
\SetMathAlphabet{\mathsfit}{bold}{\encodingdefault}{\sfdefault}{bx}{n}

\def\gE{{\mathcal{E}}}

\def\loss{{\mathcal{L}}}

\def\normal{{\mathcal{N}}}

\def\eps{{\epsilon}}

\usepackage[capitalize,noabbrev]{cleveref}

\usepackage[textsize=tiny]{todonotes}

\icmltitlerunning{Interpolating Images with Diffusion Models}

\begin{document}

\twocolumn[
\icmltitle{Interpolating between Images with Diffusion Models}



\icmlsetsymbol{equal}{*}

% \begin{icmlauthorlist}
% \icmlauthor{Clinton Wang}{mit}
% \icmlauthor{Polina Golland}{mit}
% \end{icmlauthorlist}

% \icmlaffiliation{mit}{MIT CSAIL, Cambridge, USA}

\begin{icmlauthorlist}
\icmlauthor{Clinton J. Wang and Polina Golland}{} \\
MIT CSAIL
\end{icmlauthorlist}

% \icmlcorrespondingauthor{Clinton Wang}{clintonw@csail.mit.edu}

\icmlkeywords{Latent diffusion models, image interpolation, image editing, denoising diffusion model, video generation}

{%
\begin{center}
\centering
\captionsetup{type=figure}
\includegraphics[width=\linewidth]{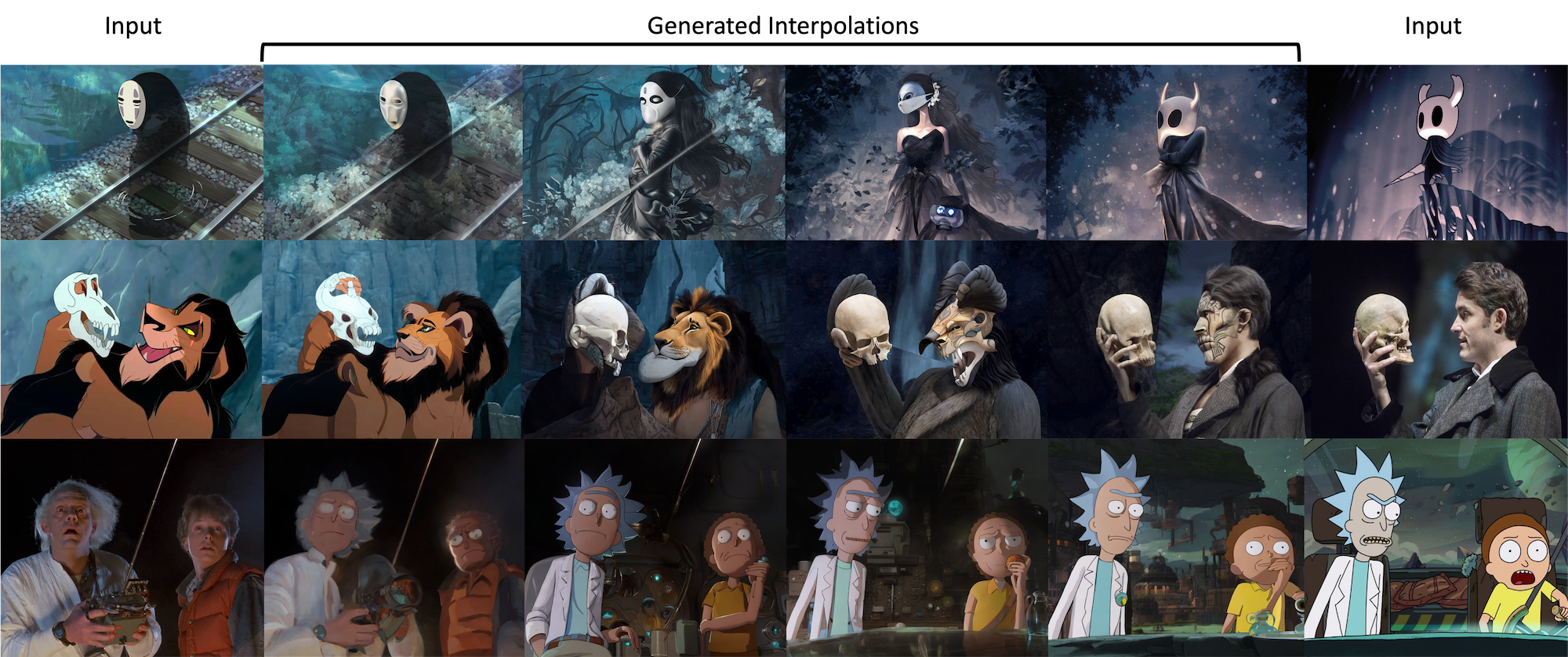}
\vspace{-12pt}
\captionof{figure}{\textbf{Interpolations of real images.}
By conditioning a pre-trained latent diffusion model on various attributes, we can interpolate pairs of images with diverse styles, layouts, and subjects.}
\label{fig:teaser}
\end{center}}

\vskip 0.2in
]



\printAffiliationsAndNotice{}  % leave blank if no need to mention equal contribution

% \input{figs/1_teaser}

\begin{abstract}
One little-explored frontier of image generation and editing is the task of interpolating between two input images, a feature missing from all currently deployed image generation pipelines. We argue that such a feature can expand the creative applications of such models, and propose a method for zero-shot interpolation using latent diffusion models. We apply interpolation in the latent space at a sequence of decreasing noise levels, then perform denoising conditioned on interpolated text embeddings derived from textual inversion and (optionally) subject poses. For greater consistency, or to specify additional criteria, we can generate several candidates and use CLIP to select the highest quality image. We obtain convincing interpolations across diverse subject poses, image styles, and image content, and show that standard quantitative metrics such as FID are insufficient to measure the quality of an interpolation. Code and data are available at \url{https://clintonjwang.github.io/interpolation}.
\end{abstract}
\vspace{-3pt}
\section{Introduction}
\label{sec:intro}

Image editing has long been a central topic in computer vision and generative modeling. Advances in generative models have enabled increasingly sophisticated techniques for controlled editing of real images \cite{imagic, controlnet, mokady2022nulltext}, with many of the latest developments emerging from denoising diffusion models \cite{ddpm, ddim, ldm, dalle2, imagen}. But to our knowledge, no techniques have been demonstrated to date for generating high quality interpolations between real images that differ in style and/or content.

Current image interpolation techniques operate in limited contexts. Interpolation between generated images has been used to study the characteristics of the latent space in generative adversarial networks \cite{stylegan, stylegan2}, but such interpolations are difficult to extend to arbitrary real images as such models only effectively represent a subset of the image manifold (\eg, photorealistic human faces) and poorly reconstruct most real images \cite{xia2022gan}. Video interpolation techniques are not designed to smoothly interpolate between images that differ in style; style transfer techniques are not designed to simultaneously transfer style and content gradually over many frames. We argue that the task of interpolating images with large differences in appearance, though rarely observed in the real world and hence difficult to evaluate, will enable many creative applications in art, media and design.

We introduce a method for using pre-trained latent diffusion models to generate high-quality interpolations between images from a wide range of domains and layouts (Fig. \ref{fig:teaser}), optionally guided by pose estimation and CLIP scoring. Our pipeline is readily deployable as it offers significant user control via text conditioning, noise scheduling, and the option to manually select among generated candidates, while requiring little to no hyperparameter tuning between different pairs of input images. We compare various interpolation schemes and present qualitative results for a diverse set of image pairs. We plan to deploy this tool as an add-on to the existing Stable Diffusion \cite{ldm} pipeline.
% However, the quality of such interpolations degenerates substantially.
% generative adversarial networks cannot obtain perfect reconstructions of real images from latent space.
% While latent diffusion models enable near-perfect inversion of real images, a naive latent space interpolation often yields poor quality intermediate frames. Our strategy is to guide the interpolation process using the information already available in the given images to yield superior interpolations. Due to the flexibility of diffusion models' conditioning information, 

% Our contributions are as follows:
% \begin{itemize}%[leftmargin=1.4em]
%     \item We introduce the \modelname, the first general framework for learning from implicit neural datasets of arbitrary parameterization.
% \end{itemize}

% \input{tables/method}
\vspace{-3pt}
\section{Related Work} \label{sec:related}

\paragraph{Image editing with latent diffusion models}
Denoising diffusion models \cite{ddpm} and latent diffusion models \cite{ldm} are powerful models for text-conditioned image generation across a wide range of domains and styles. They have become popular for their highly photorealistic outputs, degree of control offered via detailed text prompts, and ability to generalize to out-of-distribution prompts \cite{dalle2, imagen}. Follow-up research continued to expand their capabilities, including numerous techniques for editing real images \cite{imagic, brooks2023instructpix2pix, mokady2022nulltext} and providing new types of conditioning mechanisms \cite{controlnet}.

% \paragraph{Latent space interpolation}
Perhaps the most sophisticated techniques for traversing latent space have been designed in the context of generative adversarial networks (GANs), where disentanglement between style and content \cite{stylegan2}, alias-free interpolations \cite{stylegan3}, and interpretable directions \cite{lucy_steerability} have been developed. However, most such GANs with rich latent spaces exhibit poor reconstruction ability on real images, a problem referred to as GAN inversion \cite{xia2022gan}. Moreover, compared to denoising diffusion models, GANs have fewer robust mechanisms for conditioning on other information such as text or pose. Latent diffusion models such as Stable Diffusion \cite{ldm} can readily produce interpolations of generated images \cite{latentblending}, although to our knowledge this is the first work to interpolate real images in the latent space.

% \paragraph{Style transfer}
% Many works have investigated how to generate an image that combines elements of two real images, usually by incorporating the content of one image and style of another. The separation between content and style is often built into the feature space of many generative models and image encoders, allowing for the generation of a combined image by combining and manipulating the features of each image \cite{gatys2016image, li2018closedform}. Our task is distinct from this setting as we want to produce a series of interpolations that may change simultaneously in both style and content.

% \paragraph{Frame interpolation}
% Interpolation of real images emerges in the context of video frame interpolation or novel view synthesis, but in almost all cases, the basic assumption is that the observed frames mostly depict the same style and content. As a result the task often centers on inferring occlusions and reproducing motion via techniques such as optical flow prediction and depth prediction \cite{bao2019depthaware, kong2022ifrnet}.

\vspace{-3pt}
\section{Preliminaries}

Let $x$ be a real image. A latent diffusion model (LDM) consists of an encoder $\gE: x \mapsto z_0$, decoder $\mathcal{D}: z_0 \mapsto \hat{x}$, and a denoising U-Net $\eps_\theta: (z_t; t, c_{\rm{text}}, c_{\rm{pose}}) \mapsto \hat{\eps}$. The timestep $t$ indexes a diffusion process, in which latent vectors $z_0$ derived from real images are mapped to a Gaussian distribution $z_T \sim \normal(0, I)$ by composing small amounts of i.i.d. noise at each step. Each noisy latent vector $z_t$ can be related to the original input as $z_t = \alpha_t z_0 + \sigma_t \eps$, $\eps \sim \mathcal{N}(0,I)$, for parameters $\alpha_t$ and $\sigma_t$. The role of the denoising U-Net is to estimate $\eps$ \cite{ddpm}. An LDM performs gradual denoising over several iterations, producing high quality outputs that faithfully incorporate conditioning information. $c_{\rm{text}}$ is text that describes the desired image (optionally including a negative prompt), and $c_{\rm{pose}}$ represents an optional conditioning pose for human or anthropomorphic subjects. The mechanics of text conditioning is described in \cite{ldm}, and pose conditioning is described in \cite{controlnet}.

% Pre-trained latent diffusion models like Stable Diffusion can produce high-quality reconstructions, $\norm{E(D(x)) - x}$ is small for a very wide range of images $x$.

\vspace{-3pt}
\section{Real Image Interpolation}
% Let $x^0, x^N$ be two real images that we want to interpolate with $N-1$ intermediate images.

\begin{figure}[t]
\centering
\includegraphics[width=\linewidth]{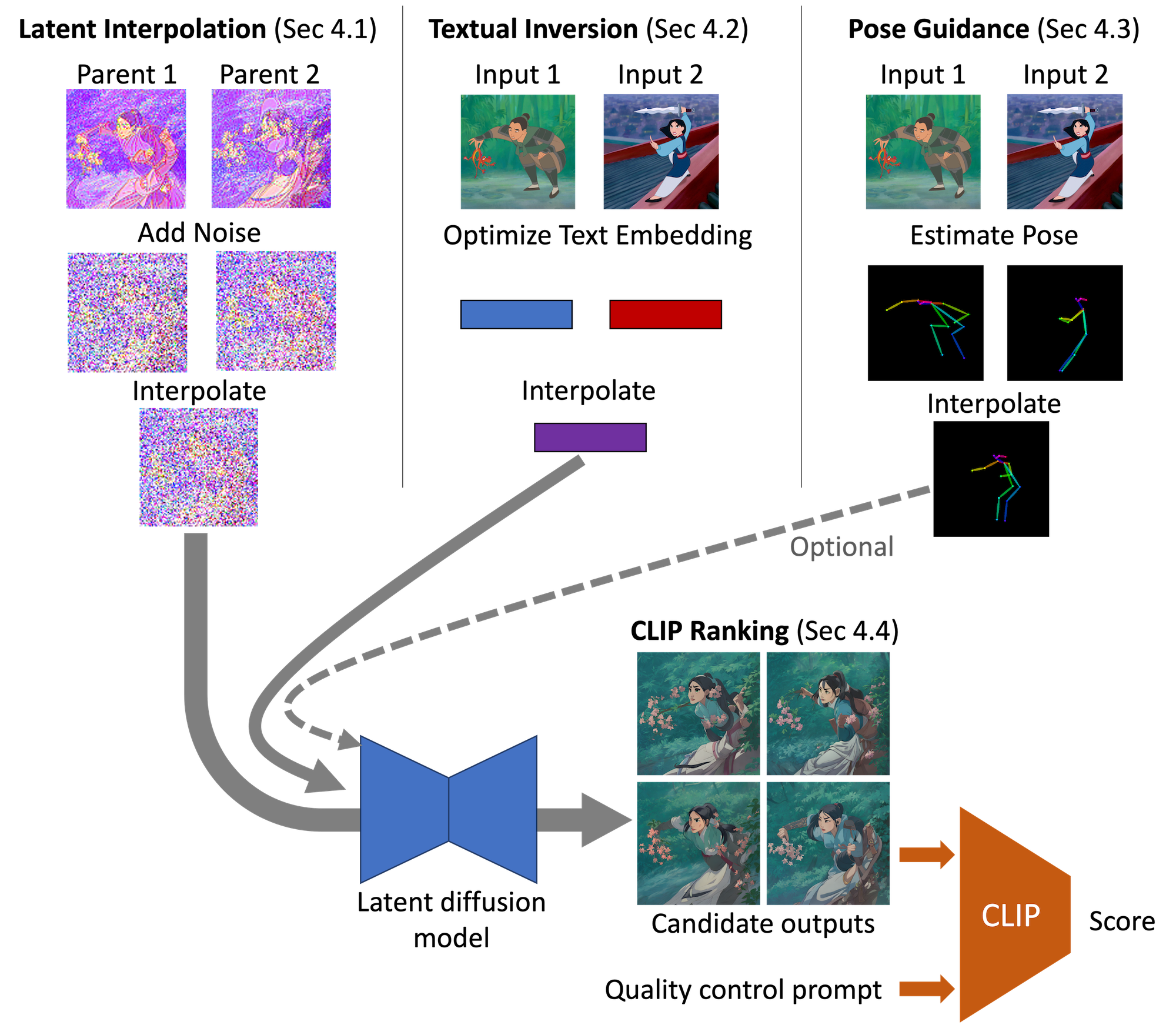}
\vspace{-5pt}
\caption{\textbf{Our pipeline.} To generate a new frame, we interpolate the noisy latent images of two existing frames (Section \ref{sec:latent_interp}). Text prompts and (if applicable) poses are extracted from the original input images, and interpolated to provide to the denoiser as conditioning inputs (Section \ref{sec:text_inversion} and \ref{sec:pose_guidance}). This process can be repeated for different noise vectors to generate multiple candidates. The best candidate is selected by computing its CLIP similarity to a prompt describing desired characteristics (Section \ref{sec:clip_ranking}).}
\label{fig:pipeline}
\vspace{-7pt}
\end{figure}

\subsection{Latent interpolation}\label{sec:latent_interp}
Our general strategy for generating sequences of interpolations is to iteratively interpolate pairs of images, starting with the two given input images. 
For each pair of parent images, we add shared noise to their latent vectors, interpolate them, then denoise the result to generate an intermediate image. The amount of noise to add to the parent latent vectors should be small if the parents are close to each other in the sequence, to encourage smooth interpolations. If the parents are far apart, the amount of noise should be larger to allow the LDM to explore nearby trajectories in latent space that have higher probability and better match other conditioning information.

Concretely, we specify a sequence of increasing timesteps $\mathcal{T}=(t_1,\dots,t_K)$, and assign parent images using the following branching structure: images $0$ and $N$ (the input images) are diffused to timestep $t_K$ and averaged to generate image $\frac{N}{2}$, images $0$ and $\frac{N}{2}$ are diffused to timestep $t_{K-1}$ generate image $\frac{N}{4}$, images $\frac{N}{2}$ and $N$ are also diffused to timestep $t_{K-1}$ to generate image $\frac{3N}{4}$, and so on. By adding noise separately to each pair of parent images, this scheme encourages images to be close to their parents, but disentangles sibling images.  %images $kN/2^j$ and $(k+2)N/2^j$ generate image $(2k+1)N/2^{j+1}$ for all $j=$

\paragraph{Interpolation type}
We use spherical linear interpolations (\textit{slerp}) for latent space and text embedding interpolations, and linear interpolations for pose interpolations. Empirically, the difference between \textit{slerp} and linear interpolation appears to be fairly mild.

\paragraph{Noise schedule}
We perform DDIM sampling \cite{ddim}, and find that the LDM's quality is more consistent when the diffusion process is partitioned into at least 200 timesteps, and noticeably degrades at coarser schedules. Empirically, latent vectors denoised with less than 25\% of the schedule often resemble an alpha composite of their parent images, while images generated with more than 65\% of the schedule can deviate significantly from their parent images. For each interpolation we choose a linear noise schedule within this range, depending on the amount of variation desired in the output. Our approach is compatible with various stochastic samplers \cite{karras2022elucidating} which seem to yield comparable results.
% Written out explicitly, we create sequences of corresponding noisy latents $\{z_t^0\}_{t \in \mathcal{T}}, \{z_t^N\}_{t \in \mathcal{T}}$, such that:
% \begin{gather}
% z_t^i = \alpha_t z_{t-1}^i + \beta_t \eps_t,
% \end{gather}
% where $\eps_t \sim \normal(0,I)$ is shared for both images.% and $z_0^0, z_0^N$ are obtained as before.
% Each intermediate image is assigned a particular timestep $t := \texttt{image_schedule}(i)$ to generate its interpolated latent code:
% $z_t^i := \texttt{slerp}(z_t^0, z_t^N, i/N)$
% We then perform denoising with the LDM: $z_0^i := \mu_\theta(z_t^i, t)$ and use the decoder to produce the image.

% $z_0^0 := \gE(x^0)$, $z_0^N := \gE(x^N)$, and all images are generated $z_0^i = \texttt{slerp}(z_0^0, z_0^N, i/N)$, $x^i := \mathcal{D}(z_0^i)$
% We examine three different strategies for latent interpolation, which differ in how they combine diffusion with interpolation to create interpolated images.

% \paragraph{Denoise-renoise-interpolate}
% Rather than partially denoise each latent, we can fully denoise the latent, then add new noise back to the appropriate level before interpolating it. This strategy permits a much wider range of latent space to be traversed, by decoupling images $N/4$ from $3N/4$, etc., while still forcing adjacent images to be similar.
% \footnote{The interpolation of two latent vectors at a particular noise level may not remain at the same noise level due to correlations introduced during the denoising process. However, we observe empirically that the independent noise assumption.}

\subsection{Textual inversion}\label{sec:text_inversion}
Pre-trained latent diffusion models are heavily dependent on text conditioning to yield high quality outputs of a particular style. Given an initial text prompt describing the overall content and/or style of each image, we can adapt its embedding more specifically to the image by applying textual inversion. In particular, we encode the text prompt as usual, then fine-tune the prompt embedding to minimize the error of the LDM on denoising the latent vector at random noise levels when conditioned on this embedding. Specifically, we perform 100-500 iterations of gradient descent with the loss $\loss(c_{\rm{text}}) = \norm{\hat{\eps}_\theta(\alpha_t z_0 + \sigma_t \eps; t, c_{\rm{text}}) - \eps}$ and a learning rate of $10^{-4}$. The number of iterations can be increased for images with complicated layouts or styles which are harder to represent with a text prompt.

In this paper we specify the same initial prompt for both input images, although one can also substitute a captioning model for a fully automated approach. Both positive and negative text prompts are used and optimized, and we share the negative prompt for each pair of images. Since our task does not require a custom token, we choose to optimize the entire text embedding.
% We also want to interpolate the prompt between the input images so that the style and content can transition smoothly. We can either specify prompts for each of the images, or perform single-image textual inversion on the images. In our experience, the best approach is to choose an initial shared positive and negative prompt for the images, then .

\subsection{Pose guidance}\label{sec:pose_guidance}
\begin{figure}[th]
\centering
\includegraphics[width=\linewidth]{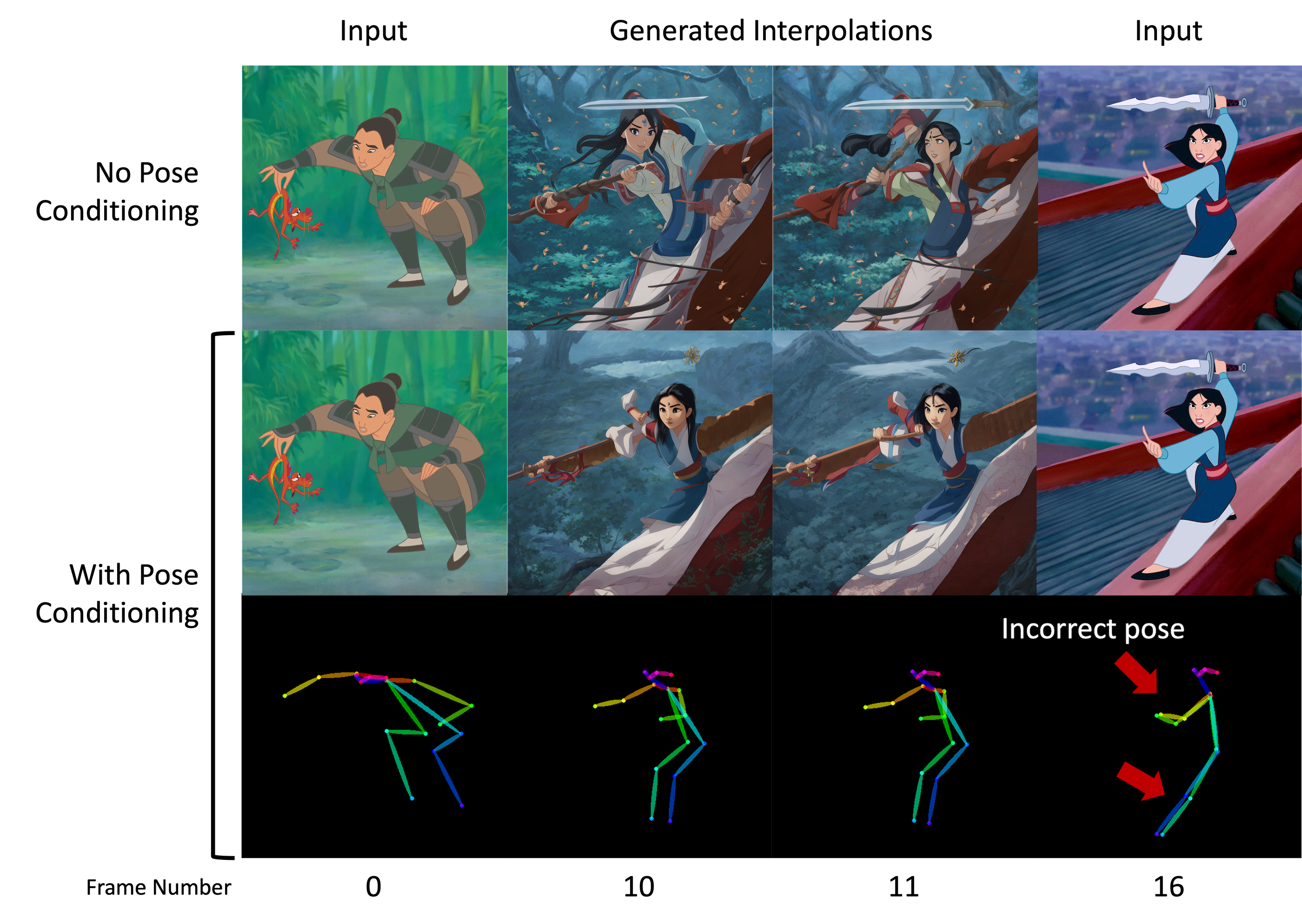}
\vspace{-10pt}
\caption{\textbf{Pose conditioning} mitigates the occurrence of abrupt pose changes between adjacent frames, even when the predicted pose is incorrect.}
\label{fig:pose}
\vspace{-5pt}
\end{figure}

\begin{figure}[ht]
% \vspace{-10pt}
\centering
\includegraphics[width=\linewidth]{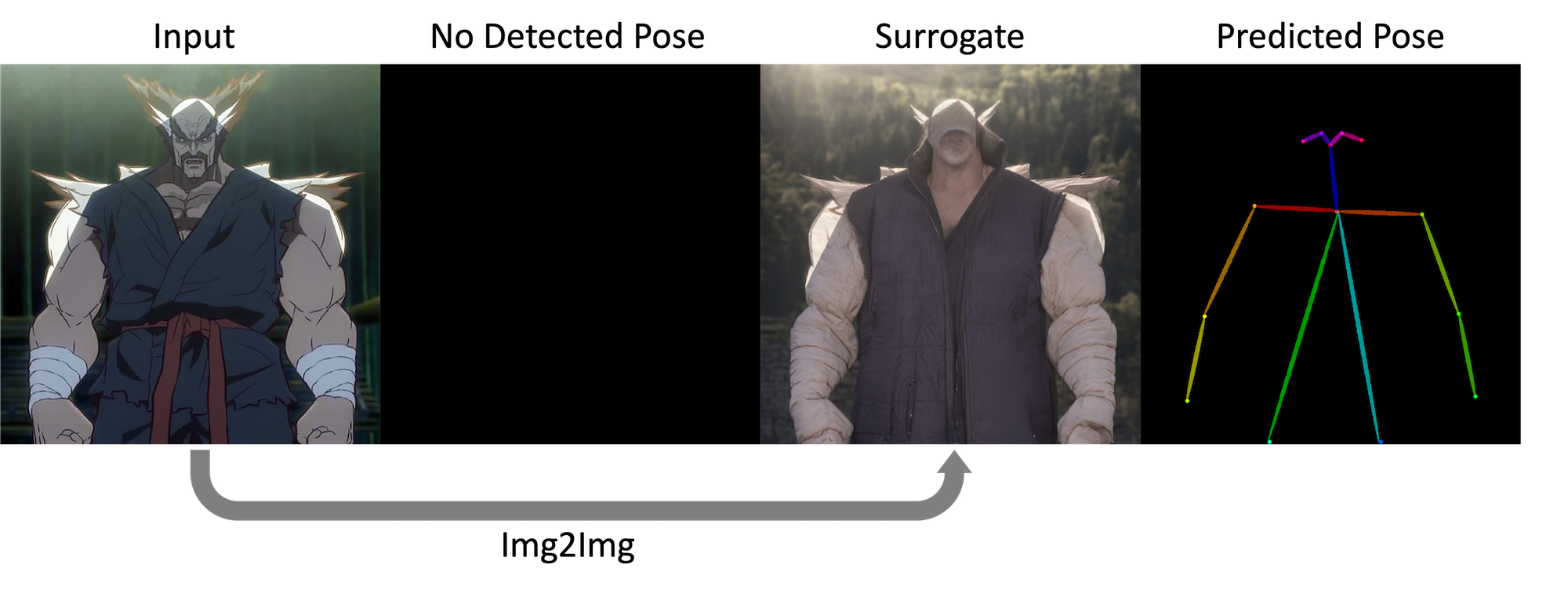}
\vspace{-12pt}
\caption{When the input image is stylized, OpenPose fails to produce a pose with high confidence. Thus we first perform image-to-image translation using our LDM, to convert the input image to the style of a photograph before applying OpenPose. It often still succeeds even when the translated image is of low quality.}
\label{fig:openpose}
\end{figure}

If the subject's pose differs significantly between the two images, image interpolation is challenging and often results in anatomical errors such as multiple limbs and faces. We obtain more plausible transitions between subjects in different poses by incorporating pose conditioning information in the LDM. We obtain poses of the input images using OpenPose \cite{openpose}, with the assistance of style transfer for cartoons or non-human subjects (see Fig. \ref{fig:openpose}). We then linearly interpolate all shared keypoint positions from the two images to obtain intermediate poses for each image. The resulting pose is provided to the LDM using ControlNet \cite{controlnet}, a powerful method for conditioning on arbitrary image-like inputs. Interestingly, we observe that \textit{even when the wrong pose is predicted} for input images, conditioning on pose still yields superior interpolations as it prevents abrupt pose changes (see Fig. \ref{fig:pose}). %Additionally, we find it helpful to increase the strength of the pose conditioning for images towards the middle of the sequence.
% If the modality of the input images is unsuitable for obtaining accurate pose information (e.g. stylized cartoons), we can first perform style transfer to a photorealistic image using the LDM, which will be more suitable as input to OpenPose even if the image quality is poor .
%, so by choosing the LDM to use 10-20 timesteps, this step becomes very fast

\vspace{-3pt}
\subsection{CLIP ranking}\label{sec:clip_ranking}
\vspace{-3pt}
LDMs can yield outputs of widely varying quality and characteristics with different random seeds. This problem is compounded in real image interpolation since a single bad generated image compromises the quality of all other images derived from it.
Thus when quality is more important than speed, multiple candidates can be generated with different random seeds, then ranked with CLIP \cite{clip}. We repeat each forward diffusion step with different noise vectors, denoise each of the interpolated latent vectors, then measure the CLIP similarity of the decoded image with specified positive and negative prompts (e.g., positive: ``high quality, detailed, 2D'', negative: ``blurry, distorted, 3D render''). The image with the highest value of positive similarity minus negative similarity is kept. %We generate more candidates for images at higher noise levels since these images have more freedom to deviate from the desired style.
In applications requiring an even higher degree of control and quality, this pipeline can be changed into an interactive mode where users can manually select desired interpolations or even specify a new prompt or pose for a particular image.

\vspace{-3pt}
\section{Experiments}\label{analysis}
\vspace{-3pt}

We analyze the effect of various design choices when applying Stable Diffusion v2.1 \cite{ldm} with pose-conditioned ControlNet on a curated set of \numpairs pairs of images spanning diverse domains (see Fig. \ref{fig:more_interps}-\ref{fig:last_interps} for more examples). They include photographs, logos and user interfaces, artwork, ads and posters, cartoons, and video games. %We include examples with landscapes, buildings, and multiple subjects., and the full set of inputs and interpolations is available at \url{masked_url}

\subsection{Latent Interpolation}

We compare our approach for latent vector interpolation against several baselines: interpolating without denoising (interpolate only), interpolating between noisy versions of the input vectors (interpolate-denoise), interpolating partially denoised versions of generated latents (denoise-interpolate-denoise), and denoise-interpolate-denoise with no shared noise added to the input latents.

\paragraph{Interpolate only}
The naive interpolation scheme simply interpolates the clean latent codes of the input images without performing any diffusion. We set $z_0^0 := \gE(x^0)$, $z_0^N := \gE(x^N)$, and all images are generated via $z_0^i = \texttt{slerp}(z_0^0, z_0^N, i/N)$, $x^i := \mathcal{D}(z_0^i)$. This approach completely fails to generate reasonable images as the denoised latent space in LDMs is not well-structured. 

\paragraph{Interpolate-denoise}
We choose a sequence of increasing timesteps $\mathcal{T}=(0,\dots,T)$ and create sequences of corresponding noisy latents $\{z_t^0\}_{t \in \mathcal{T}}, \{z_t^N\}_{t \in \mathcal{T}}$, such that:
\begin{gather}
z_t^0 = \alpha_t z_{t-1}^0 + \beta_t \eps_t, \\
z_t^N = \alpha_t z_{t-1}^N + \beta_t \eps_t,
\end{gather}
where $\eps_t \sim \normal(0,I)$ is shared for both images, and $z_0^0, z_0^N$ are obtained as before.
Each intermediate image is assigned a particular timestep $t := \texttt{frame\char`_schedule}(i)$ to generate its interpolated latent code: $z_t^i := \texttt{slerp}(z_t^0, z_t^N, i/N)$. \texttt{frame\char`_schedule} is a function that monotonically decreases as its input approaches 0 or $N$, to support smooth interpolation close to the input images. We then perform denoising with the LDM: $z_0^i := \mu_\theta(z_t^i, t)$ and use the decoder to produce the image. 
% We then generate interpolated latent codes , $x^i := \mathcal{D}(z_0^i)$assign lower timesteps to frames near the beginning and end of the sequence so that they remain similar to the input images. 
% This interpolation scheme diffuses the interpolated latent codes in the sequence such that , using shared noise in the forward diffusion process.

\paragraph{Denoise-interpolate-denoise}
If we rely on $\{z_t^0\}$ and $\{z_t^N\}$ to generate all intermediate latents, adjacent images at high noise levels may diverge significantly during the denoising process. Instead, we can interpolate images in a branching pattern as follows: we first generate $z_{t_1}^{N/2}$ as an interpolation of $z_{t_1}^0$ and $z_{t_1}^N$, denoise it to time $t_2$, then generate $z_{t_2}^{N/4}$ as an interpolation of $z_{t_2}^0$ and $z_{t_2}^{N/2}$, and generate $z_{t_2}^{3N/4}$ similarly. These two new latents can be denoised to time $t_3$, and so on. The branching factor can be modified at any level so the total number of frames does not need to be a power of 2. This interpolation scheme is similar to latent blending \cite{latentblending}.

% is visualized in Fig. \ref{fig:interp} and
% Because the latent space traversed between $\{z_t^0\}$ and $\{z_t^N\}$ is not necessarily well-behaved.
% This interpolation scheme performs interpolation of noisy latent codes. We perform forward diffusion on the latents of the input images using shared noise. We first generate the midpoint latent by interpolating these noisy latents. We reverse diffuse this latent for some number of steps, then generate quartile latents by interpolating the partially denoised midpoint latent with the input latents at the same noise level. This interpolation scheme is visualized in Fig. \ref{fig:interp}.

\begin{figure}[th]
\centering
\includegraphics[width=\linewidth]{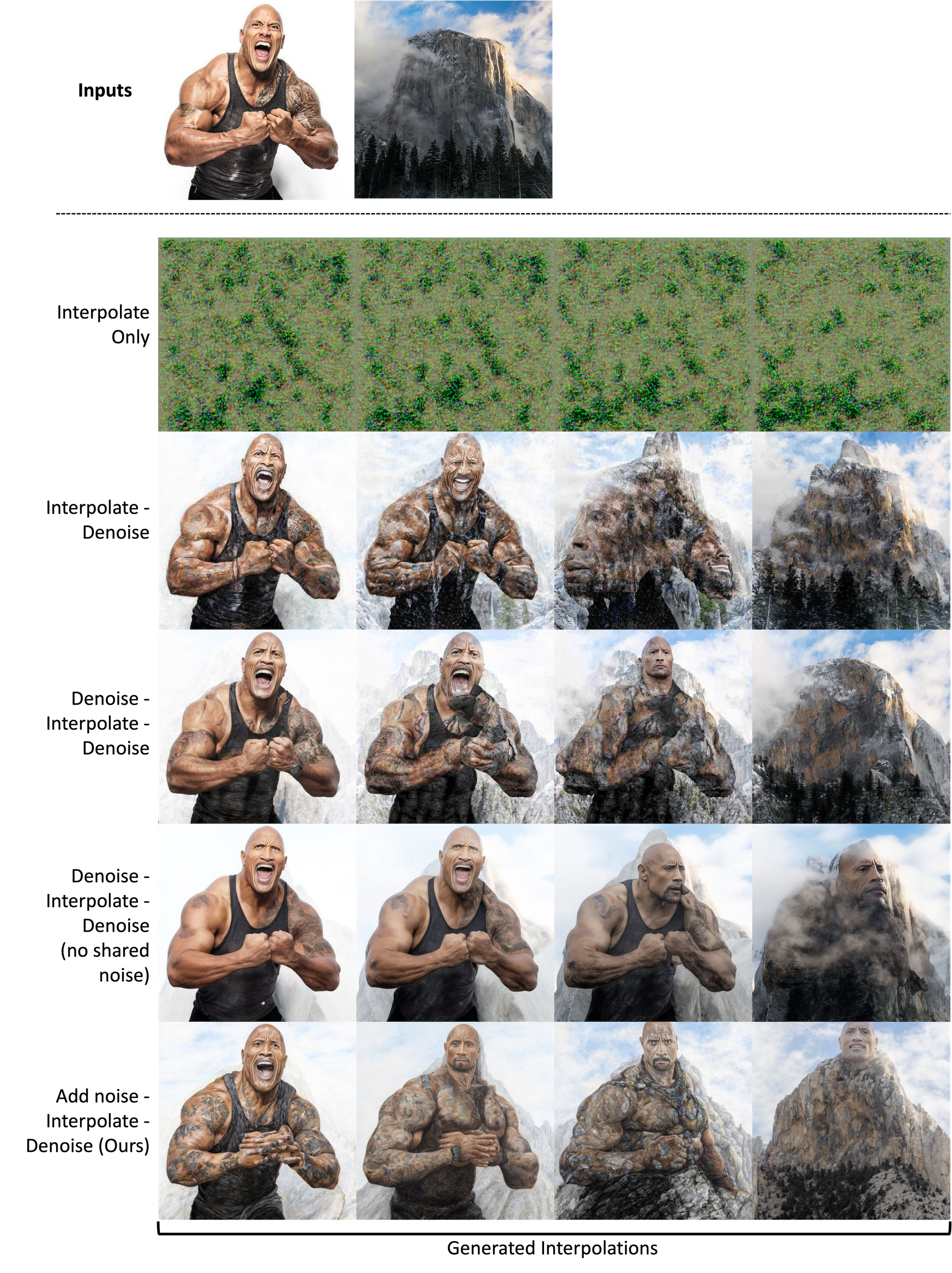}
\vspace{-13pt}
\caption{\textbf{Comparison of different interpolation schemes.}
We add noise to the latents derived from our input images, and denoise the interpolated latents to generate output frames. This approach performs a more convincing semantic transformation from a human to a mountain compared to other approaches which instead resemble alpha blending.}
\label{fig:comparison}
\vspace{-5pt}
\end{figure}

Qualitatively we found that the most convincing and interesting interpolations were achieved by our method (Fig. \ref{fig:comparison}). Other interpolation schemes either fully couple the noise between all frames, which results in less creative outputs that resemble alpha blending rather than a semantic transformation, or do not perform any noise coupling, which can result in abrupt changes between adjacent frames. 
Interestingly this phenomenon is not captured by distributional metrics such as Fr\'echet inception distance (FID) \cite{fid} or smoothness metrics such as perceptual path length (PPL) \cite{stylegan2} (see Table \ref{tbl:performance}). We computed the FID between the distribution of input images and distribution of output images (two random frames sampled from every interpolation) as a proxy for the degree to which output images lie on the image manifold. We compute PPL as the sum of Inception v3 distances between adjacent images in 17-frame sequences, to measure the smoothness of the interpolations and the degree to which the interpolation adheres to the appearance of the input images. We find that both these metrics favor interpolations that resemble simple alpha composites rather than more creative interpolations, as the latter deviate more in feature statistics from the original set of images, even if they would be preferred by users. Thus current metrics are insufficient to capture the effectiveness of an interpolation, an open question that we hope to tackle in future work.

\begin{table}[th]
\centering
% \vspace{-5pt}
\caption{\textbf{Quantitative comparison.} Fr\'echet inception distance (FID) between input images and their interpolations, and perceptual path length (PPL, mean$\pm$std) of each interpolation in Inception v3 feature space.}\label{tbl:performance}
% \vspace{-5pt}
\begin{tabular}{lcc}
\hline
\textbf{Interpolation Scheme} & \textbf{FID} & \textbf{PPL}\\
\hline
Interpolate only & 436 & 56$\pm$8 \\
Interpolate-denoise & 179 & 172$\pm$32 \\
Denoise-interpolate-denoise (DID) & 169 & 144$\pm$26 \\
DID w/o shared noise & 199 & 133$\pm$22 \\
Add noise-interpolate-denoise (ours) & 214 & 193$\pm$27 \\
\hline
\vspace{-5pt}
\end{tabular}
\end{table}

\vspace{-3pt}
\subsection{Extensions}
% The original formulation of diffuse then interpolate uses the same 

\paragraph{Interpolation schedule}
In all examples presented in this paper, we use a uniform interpolation schedule. But evenly spaced interpolations in the latent space do not necessarily translate to a constant rate of perceptual changes in the image space. While coloration and brightness seem to evolve at a constant rate between frames, we observe that stylistic changes can occur very rapidly close to the input images (for example, the transition from real to cartoon eyes in the third row of Fig. \ref{fig:teaser}). Thus in applications where the user would like to control the rate of particular changes, it can be helpful to specify a non-uniform interpolation schedule.
% Empirically, we find that rapid latents encoded from real images tend to be less stable, meaning that even small steps in latent space can yield images that appear very different from the original image. To counteract this effect, we schedule the interpolation coefficient so that steps in latent space are smaller close to the original images, which yields a more perceptually smooth sequence of intermediate images.

\begin{figure}[ht]
\centering
\includegraphics[width=\linewidth]{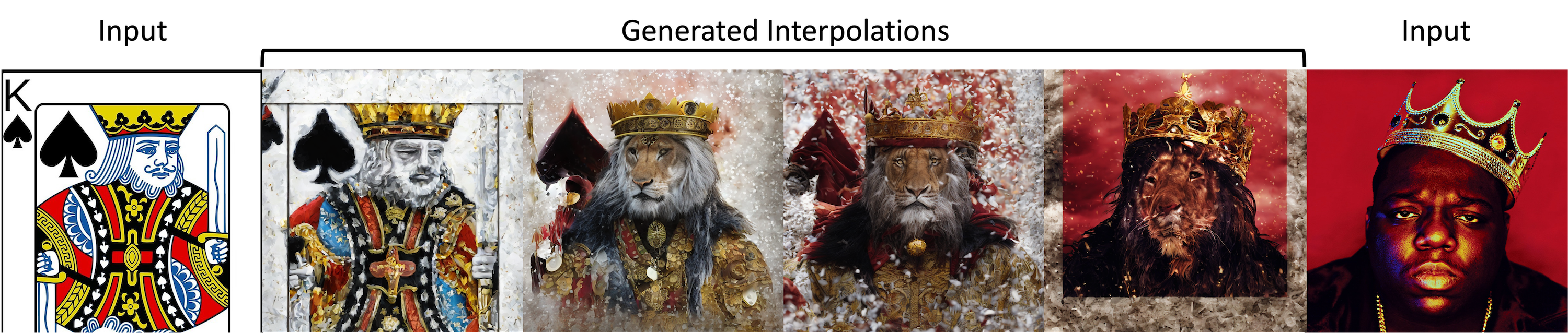}
\caption{Our pipeline can be combined with affine transforms such as zooming on a point.}
\label{fig:zoom}
\end{figure}

\paragraph{Adding motion}
Interpolation can be combined with affine transforms of the image in order to create the illusion of 2D or 3D motion (Fig. \ref{fig:zoom}). Before interpolating each pair of images, we can warp the latent of one of the images to achieve the desired transform.

\vspace{-3pt}
\section{Conclusion}
\vspace{-3pt}

We introduced a new method for real image interpolation that can generate imaginative, high-quality sequences connecting images with different styles, content and poses.
This technique is quite general, and can be readily integrated with many other methods in video and image generation such as specifying intermediate prompts, and conditioning on other inputs such as segmentations or bounding boxes.

\paragraph{Limitations}
Our method can fail to interpolate pairs of images that have large differences in style and layouts. In Fig. \ref{fig:failures}, we illustrate examples where the model cannot detect and interpolate the pose of the subject (top), fails to understand the semantic mapping between objects in the frames (middle), and struggles to produce convincing interpolations between very different styles (bottom). We also find that the model occasionally inserts spurious text, and can confuse body parts even given pose guidance.

%%%%%%%%% REFERENCES
% \clearpage
\pagebreak
% {\small
\bibliographystyle{assets/icml2023}
\bibliography{biblio}
% }

\newpage
\renewcommand\thefigure{\thesection.\arabic{figure}}
\renewcommand\thetable{\thesection.\arabic{table}}
\appendix
\onecolumn
\setcounter{figure}{0}
\setcounter{table}{0}
% \setcounter{algorithm}{0}
% \textbf{\Large{Appendix}}
\section{Additional Figures}

\begin{figure*}[ht]
\centering
\vspace{-10pt}
\includegraphics[width=.9\linewidth]{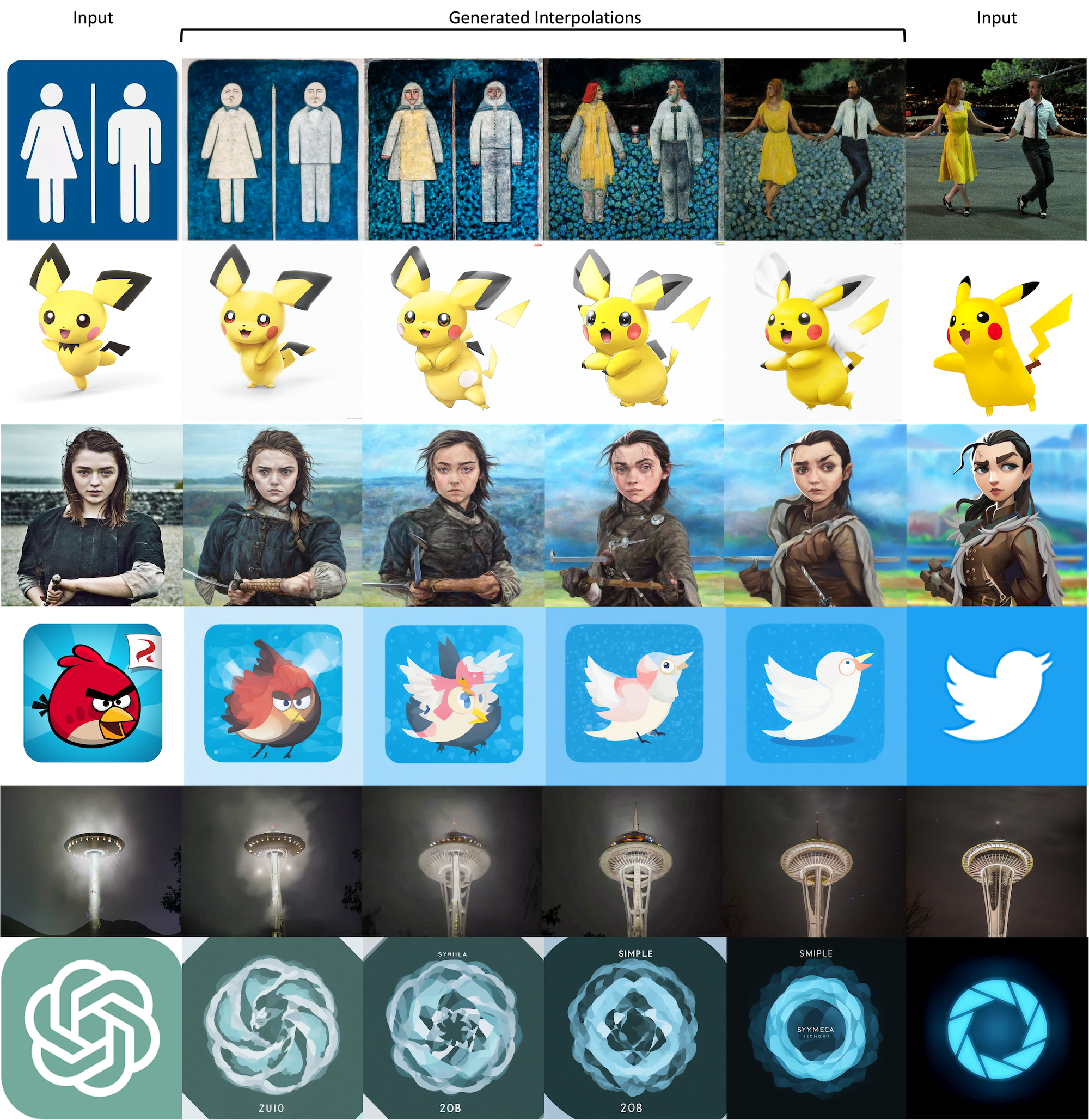}
\vspace{-10pt}
\caption{Additional image interpolations (1/3).}
\label{fig:more_interps}
\vspace{-10pt}
\end{figure*}

\begin{figure*}[ht]
\centering
\includegraphics[width=\linewidth]{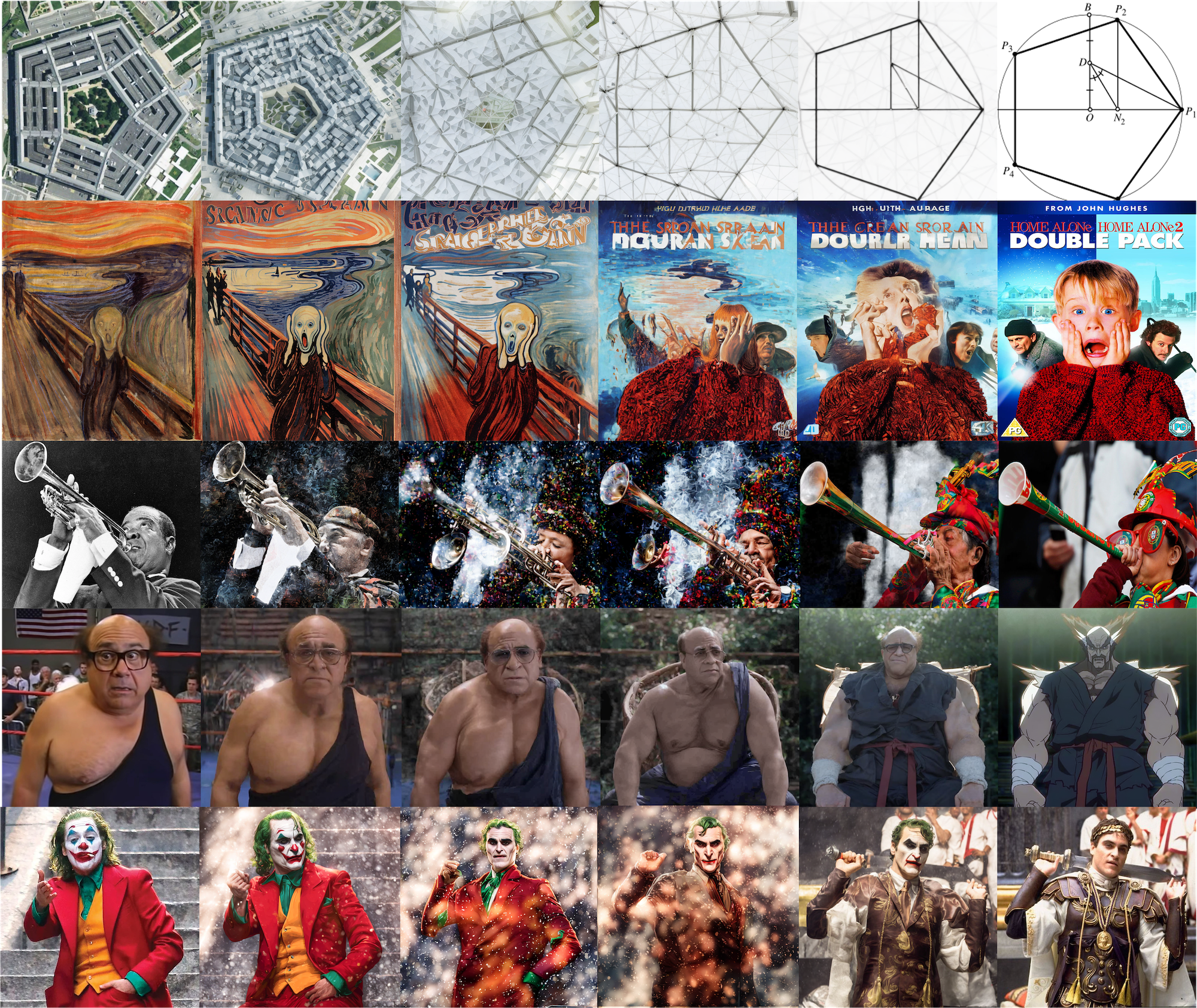}
\caption{Additional image interpolations (2/3).}
\end{figure*}

\begin{figure*}[ht]
\centering
\includegraphics[width=\linewidth]{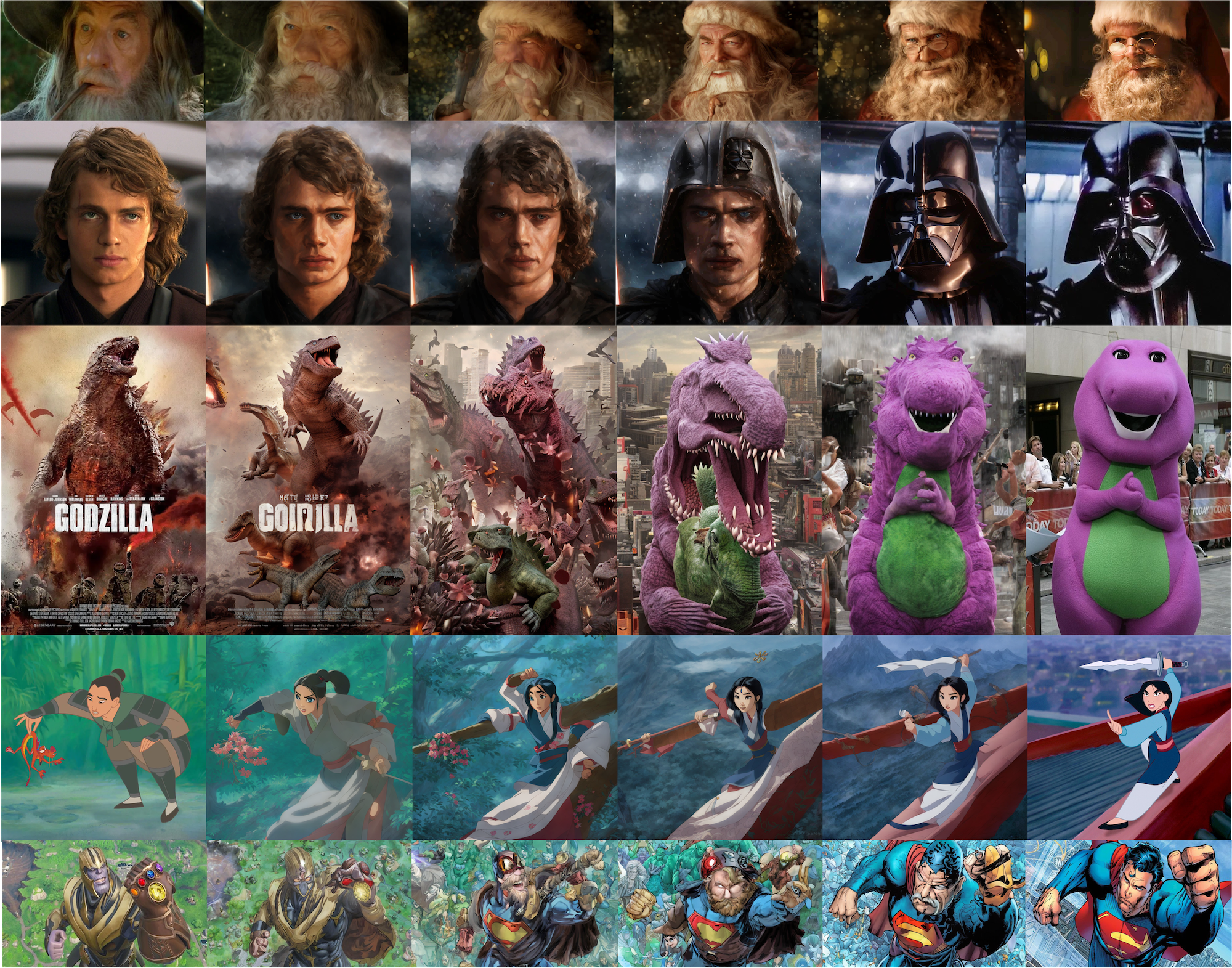}
\caption{Additional image interpolations (3/3).}
\label{fig:last_interps}
\end{figure*}

\begin{figure*}[ht]
\centering
\includegraphics[width=.7\linewidth]{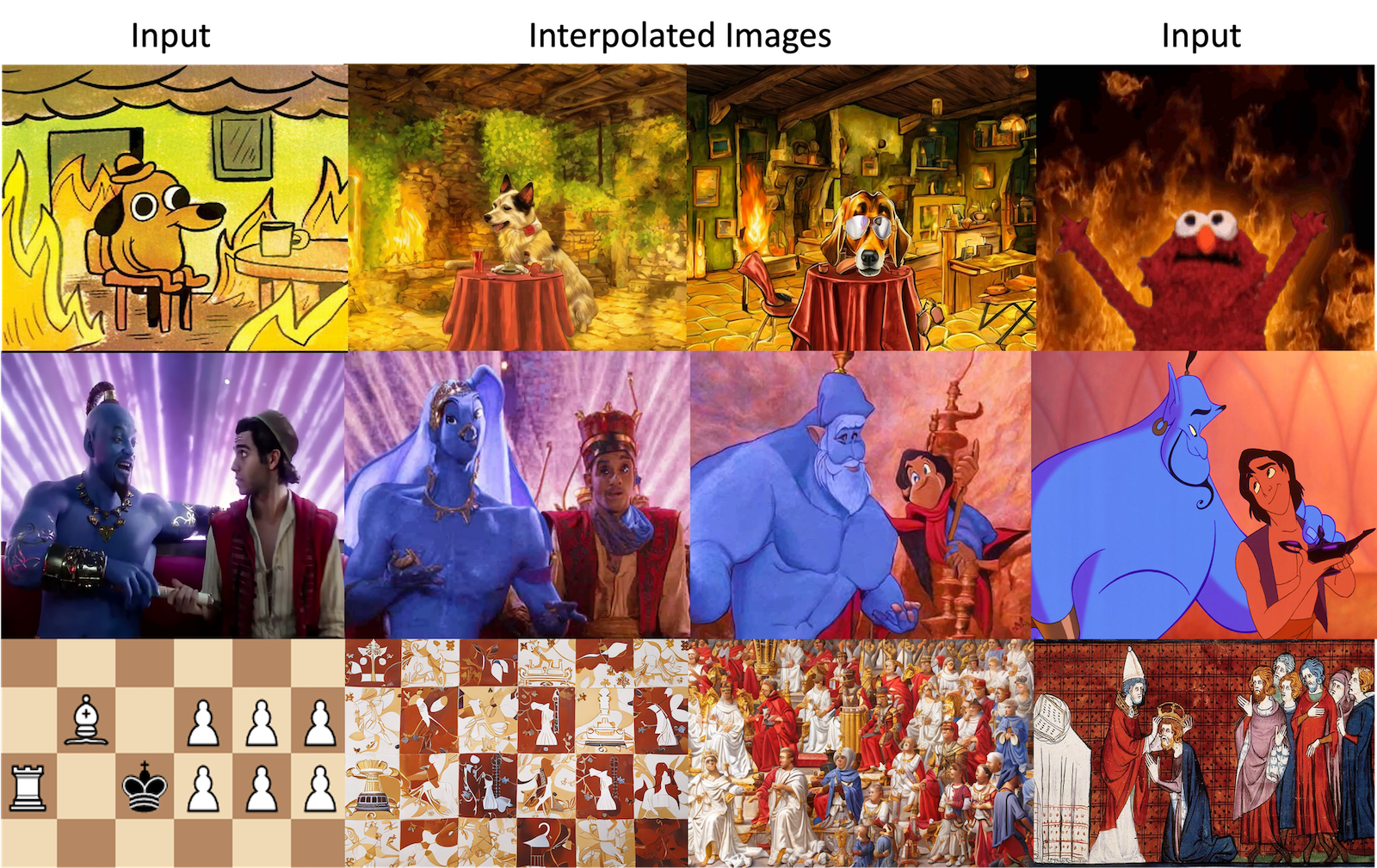}
\caption{Failure cases. Our approach is still limited in its ability to bridge large gaps in style, semantics and/or layout.}
\label{fig:failures}
\end{figure*}

\end{document}